\documentclass[runningheads]{llncs}
\usepackage{graphicx}
\usepackage{amsmath,amssymb,bm} 
\usepackage{graphicx,subfigure}
\usepackage{color}
\usepackage{floatrow}
\usepackage{array}
\usepackage[width=122mm,left=12mm,paperwidth=146mm,height=193mm,top=12mm,paperheight=217mm]{geometry}
\usepackage{xcolor,colortbl}

\newcommand{\eg}{e.g. }

\newcommand{\ie}{i.e. }

\newcolumntype{L}[1]{>{\raggedright\let\newline\\\arraybackslash\hspace{0pt}}m{#1}}
\newcolumntype{C}[1]{>{\centering\let\newline\\\arraybackslash\hspace{0pt}}m{#1}}
\newcolumntype{R}[1]{>{\raggedleft\let\newline\\\arraybackslash\hspace{0pt}}m{#1}}

\begin{document}
\setcounter{secnumdepth}{3}
\pagestyle{headings}
\mainmatter
\def\ECCV16SubNumber{xxxx}  

\title{Weakly Supervised Learning of Affordances} 

\titlerunning{Weakly Supervised Learning of Affordances}

\authorrunning{Abhilash Srikantha, Juergen Gall}

\author{Abhilash Srikantha$^{1,2}$, Juergen Gall$^{1}$}
\institute{University of Bonn, Max Planck Institute for Intelligent Systems, Tuebingen}

\maketitle

\begin{abstract}
Localizing functional regions of objects or \textit{affordances} is an important aspect of scene understanding. 
In this work, we cast the problem of affordance segmentation as that of semantic image segmentation.
In order to explore various levels of supervision, we introduce a pixel-annotated affordance dataset of 3090 images containing 9916 object instances with rich contextual information in terms of human-object interactions. 
We use a deep convolutional neural network within an expectation maximization framework to take advantage of weakly labeled data like image level annotations or keypoint annotations.  
We show that a further reduction in supervision is possible with a minimal loss in performance when human pose is used as context. 
\end{abstract}

\section{Introduction}
The capability to perceive functional aspects of an environment is highly desired because it forms the essence of devices intended for collaborative use. 
These aspects can be categorized into abstract descriptive properties called \textit{attributes}~\cite{parikh2011relative,liu2011recognizing,patterson2012sun} or physically grounded regions called \textit{affordances}.
Affordances are important as they form the key 
representation to describe potential interactions. 
For instance, 
autonomous navigation depends heavily on understanding outdoor semantics to decide if the lane is \textit{changable} or if the way ahead is \textit{drivable}~\cite{chen2015deepdriving}.
Similarly, assistive robots must have the capability of anticipating indoor semantics like which regions of the kitchen are \textit{openable} or \textit{placeable}~\cite{koppula2013learning}.
Further, because forms of interaction are fixed for virtually any object class, it is desirable to have recognition systems that are capable of localizing functionally meaningful regions or \textit{affordances} alongside contemporary object recognition systems.

In most previous works, affordance labeling has been addressed as a stand-alone task. For instance, the methods~\cite{katz2014perceiving,kim2014semantic,myers2015affordance,hermans2011affordance} learn pixel-wise affordance labels using supervised learning techniques. 
Creating pixelwise annotated datasets, however, is heavily labor intensive. 
Therefore, in order to simplify the annotation process, current affordance datasets have been captured in highly controlled environments like a turntable setting~\cite{myers2015affordance}.
This, however, does not allow to study 
contextual information, specially those of humans, which affordances are intrinsically related to. 
One of the contributions of this work is to propose a pixel annotated affordance dataset within the purview of human interactions, thus creating possibilities to tap rich contextual information thereby fostering work towards reduced supervision levels. In addition, we show that state-of-the-art end-to-end learning techniques in semantic segmentation significantly outperform state-of-the-art supervised learning methods for affordances.  

\begin{figure}[t]
\centering
\subfigure{\includegraphics[height=40mm]{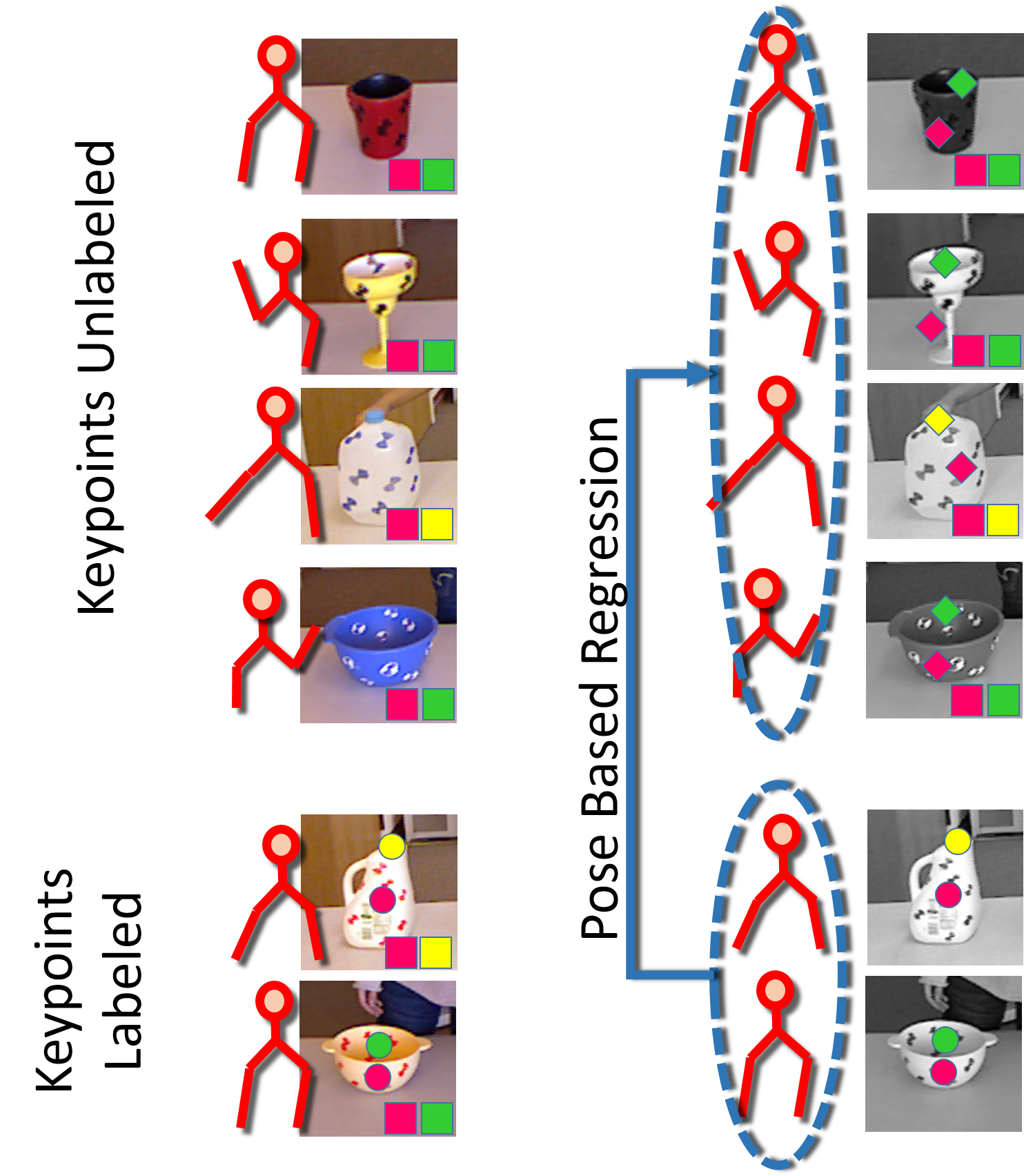}} \hspace{20mm}
\subfigure{\includegraphics[height=40mm]{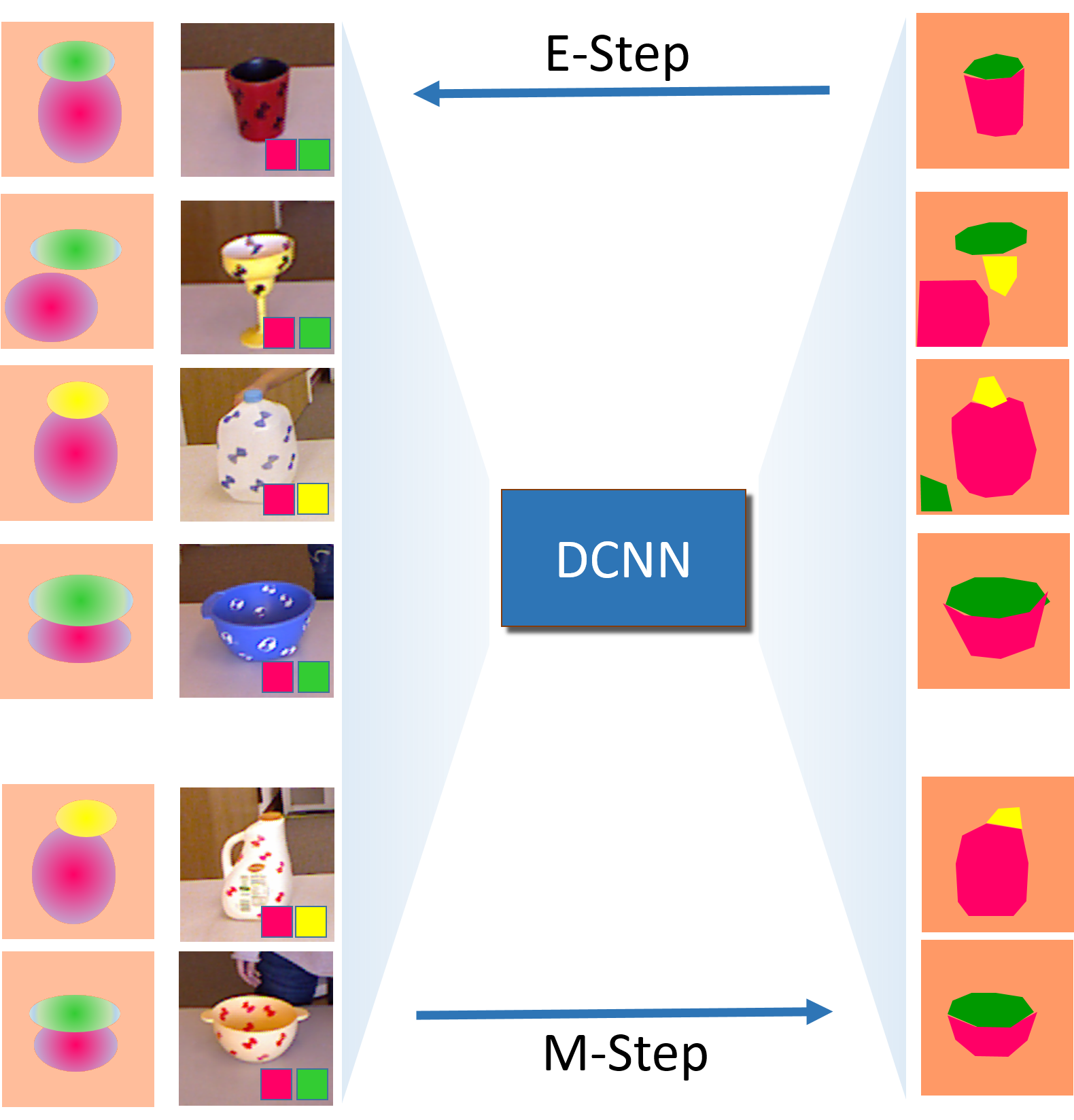}}
\caption{\label{fig:teaser} The proposed weakly supervised approach for affordance labeling. (Left) Assuming human pose and image labels are available for all images whereas keypoints as weak affordance labels are available only for a subset, we perform pose based regression to estimate keypoint locations in all other images (Section~\ref{sec:hpose_keypoint_regression}). 
(Right) The estimated keypoint locations can be used to initialize the EM framework (Section~\ref{sec:keypoints_annotation}). 
The E-step computes a point estimate of the latent segmentation based on Gaussian distributions. 
The M-step learns the parameters of the DCNN considering the point estimate as groundtruth segmentation (Section~\ref{sec:weak}). 
}
\end{figure}

As a second contribution, we propose a weakly supervised learning approach for affordance segmentation. 
Our approach is based on the expectation-maximization (EM) framework as proposed in~\cite{papandreou2015weakly}. 
The method introduces a constant bias term to learn a deep convolutional neural network (DCNN) for semantic segmentation only from image level labels. 
In this work, we consider keypoints or click-points as weak annotations, which are easy to obtain and have been used in~\cite{bell2015material} for annotating a large material database. 
In order to learn from keypoints, we extend the framework to handle spatial dependencies. 
The approach can also be used to learn from mixed sets of training images where one set is annotated by keypoints and the other is annotated by image labels. 
An overview of the proposed EM approach is illustrated in Figure~\ref{fig:teaser}.

As our third contribution, we show that automatically extracted human pose information can be effectively utilized as context for affordances. 
We use it to transfer keypoint annotations to images without keypoint annotations, which are then used to initialize the proposed EM approach.      

\section{Related Work}\label{sec:related_work}

Properties of objects can be described at various levels of abstraction by a variety of attributes including visual properties~\cite{parikh2011relative,khan2012color,farhadi2009describing,lampert2009learning} \eg object color, shape and object parts, physical properties~\cite{ferrari2007learning,zhu2014reasoning} \eg weight, size and material characteristics and categorical properties~\cite{akata2015evaluation,deng2012hedging}. 
Object affordances, which describe potential uses of an object, can also be considered as other attributes. 
For instance, \cite{chao2015mining} describes affordances by object-action pairs whose plausibility is determined either by mining word co-occurances in textual data or by measuring visual consistency in images returned by an image search.    
\cite{zhu2014reasoning} proposes to represent objects in a densely connected graph structure. 
While a node represents one of the various visual, categorical, physical or functional aspects of the object, an edge indicates the plausibility of both node entities to occur jointly. 
Upon querying the graph with observed information \eg \textit{\{round, red\}}, the result is a set of most likely nodes \eg \textit{\{tomato, edible, 10-100gm, pizza\}}. 

Affordances have also been used as an intermediate representation for higher level tasks. 
In~\cite{castellini2011using}, object functionality is defined in terms of the pose of relevant hand-grasp during interactions. 
Object recognition is performed by combining individual classifiers based on object appearance and hand pose.
\cite{zhu2015understanding} uses affordances as a part of a task oriented object modeling. 
They formulate a generative framework that encapsulates the underlying physics, functions and causality of objects being used as tools. 
Their representation combines extrinsic factors that include human pose sequences and physical forces such as velocity and pressure and intrinsic factors that represent object part affordances.
\cite{koppula2016anticipating} models action segments using CRFs which are described by human pose, object affordance and their appearances. 
Using a particle filter framework, future actions are anticipated by sampling from a pool of possible CRFs thereby performing a temporal segmentation of action labels and object affordances. \cite{kjellstrom2011visual} jointly models object appearance and hand pose during interactions. 
They demonstrate simultaneous hand action localization and object detection through implicit modeling of affordances. 

Localizing object affordances 
based on supervised learning has been popular in the robotics community. 
\cite{katz2014perceiving} performs robotic manipulations on objects based on affordances which are inferred from the orientations of object surfaces.
\cite{kim2014semantic} learns a discriminative model to perform affordance segmentation of point clouds based on surface geometry.
\cite{myers2015affordance} uses RGB-D data to learn pixelwise labeling of affordances for common household objects. 
They explore two different features: one based on a hierarchical matching pursuit and another based on normal and curvature features derived from RGB-D data.
\cite{hermans2011affordance} learns to infer object level affordance labels based on attributes derived from appearance features.
\cite{lenz2015deep} proposes a two stage cascade approach based on RGB-D data to regress potential grasp locations of objects. 
In~\cite{desai2013predicting}, pixelwise affordance labels of objects are obtained by warping the query image to the K-nearest training images based on part locations inferred using deformable part models.
\cite{song2015learning} combines top-down object pose based affordance labels with those obtained from bottom-up appearance based features to infer part-based object affordances. 
Top-down approaches for affordance labeling has been explored in~\cite{grabner2011makes,Jiang_2013_CVPR} where scenes labeling is performed by observing possible interactions between scene geometry and hallucinated human poses. 
Localizing object affordances based on human context has been also studied in~\cite{koppula2014physically}. 
They propose a graphical model where spatial and temporal extents of object affordances are inferred based on observed human pose and object locations. 
A mixture model is used to model temporal trajectories where each component represents a single type of motion \eg repetitive or random motion. The approach, however, does not provide pixelwise segmentations. Instead, the coarse location of an affordance is described by a distribution.  
Weakly-supervised learning for semantic image segmentation has been investigated in several works. In this context, training images are only annotated at the image-level and not at pixel-level. 
For instance,~\cite{vezhnevets2010towards} formulate the weakly supervised segmentation task as a multi-instance multi-task learning problem.
Further,~\cite{vezhnevets2011weakly,vezhnevets2012weakly} incorporate latent correlations among superpixels that share the same labels but originate from different images. 
\cite{xu2014tell} simplifies the above formulation by a graphical model that simultaneously encodes semantic labels of superpixels and presence or absence of labels in images.
\cite{zhang2015weakly} handle noisy labels from social images by using robust mid-level representations derived through topic modeling in a CRF framework. 
More recently, a weakly-supervised approach based on a deep DCNN~\cite{chen2014semantic} has been proposed in~\cite{papandreou2015weakly}. It uses an EM framework to iteratively learn the latent pixel labels of the training data and the parameters of the DCNN. 
A similar approach is followed by~\cite{pathak2015constrained} where linear constraints derived from weak image labels are imposed on the label prediction distribution of the neural network. 

To investigate the problem of weakly labeled affordance segmentation, we first introduce a pixel-wise labeled dataset that contains objects within the context of human-object interactions in Section~\ref{sec:datasets}. 
We then investigate various forms for weak labels and propose an EM framework that is adaptive to local image statistics in Section~\ref{sec:proposed_method}. 
In Section~\ref{sec:hpose_keypoint_regression} we show that contextual information in terms of automatically extracted human pose can be utilized to initialize the EM framework thereby further reducing the need for labeled data. 
Finally, we present evaluations in Section~\ref{sec:experiments}. 

\begin{figure}[t]
\centering
\subfigure{\includegraphics[height=10mm]{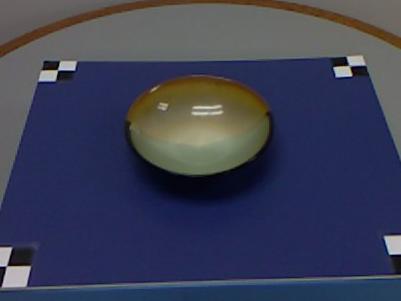}}
\subfigure{\includegraphics[height=10mm]{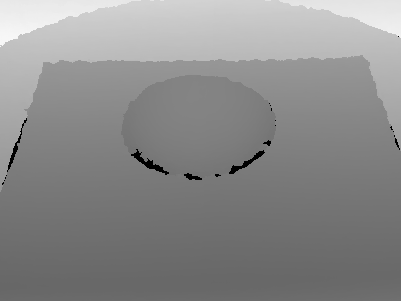}}
\subfigure{\includegraphics[height=10mm]{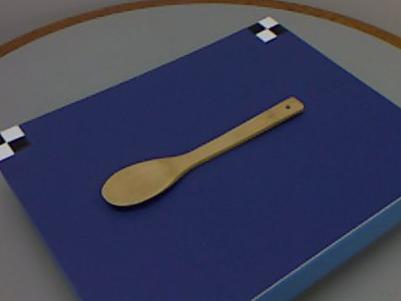}}
\subfigure{\includegraphics[height=10mm]{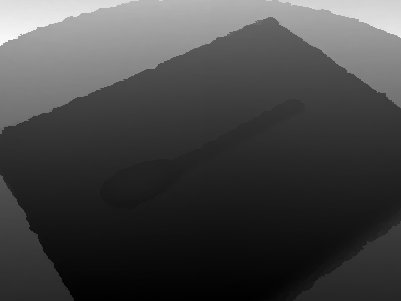}}
\subfigure{\includegraphics[height=10mm]{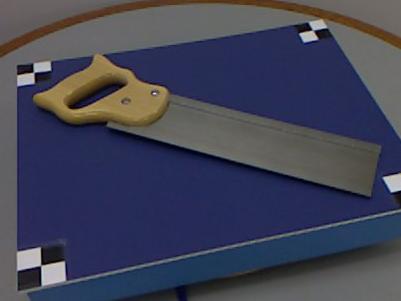}}
\subfigure{\includegraphics[height=10mm]{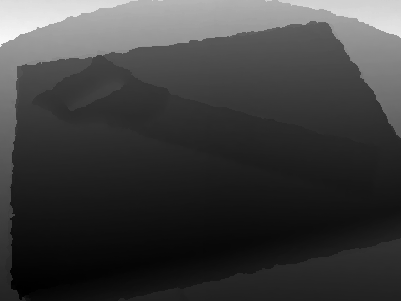}} \hspace{5mm}
\subfigure{\includegraphics[height=10mm]{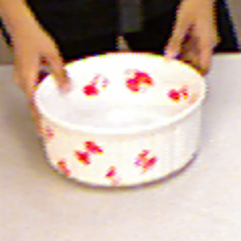}}
\subfigure{\includegraphics[height=10mm]{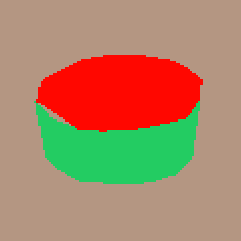}} 
\subfigure{\includegraphics[height=10mm]{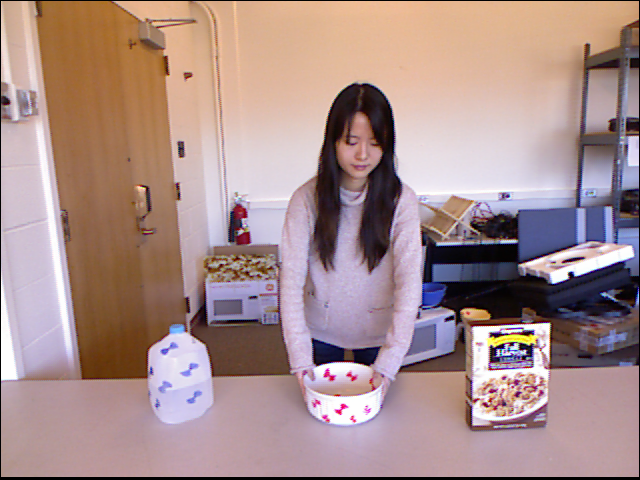}} 
\subfigure{\includegraphics[height=10mm]{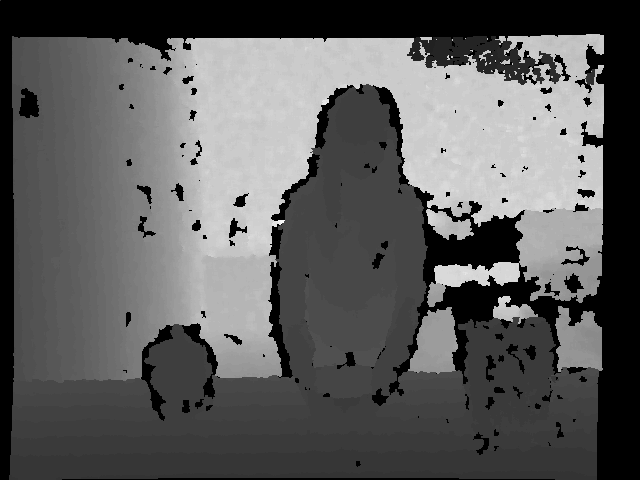}}
\subfigure{\includegraphics[height=10mm]{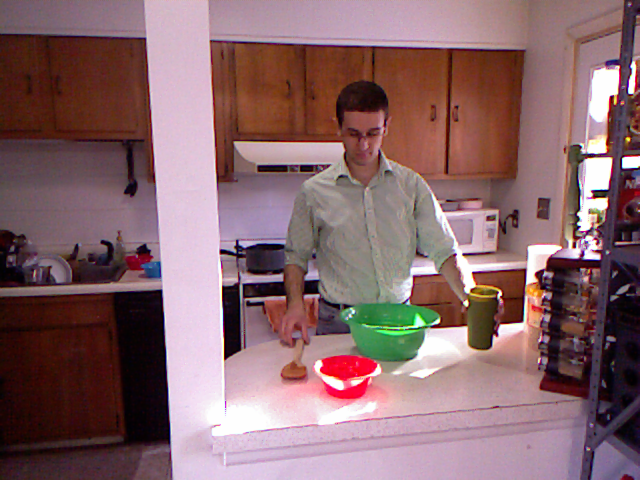}}
\subfigure{\includegraphics[height=10mm]{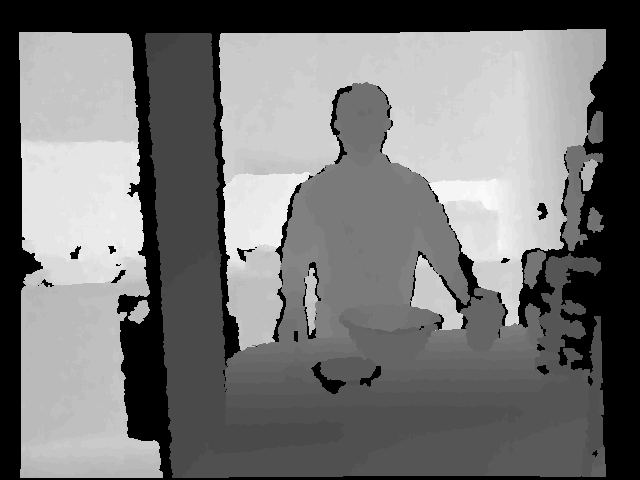}}
\subfigure{\includegraphics[height=10mm]{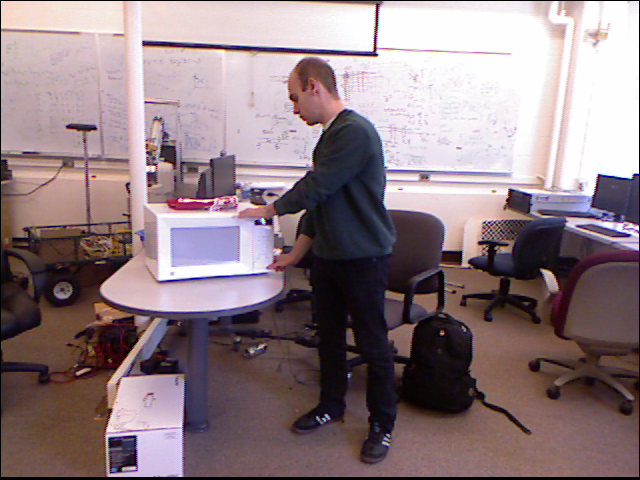}}
\subfigure{\includegraphics[height=10mm]{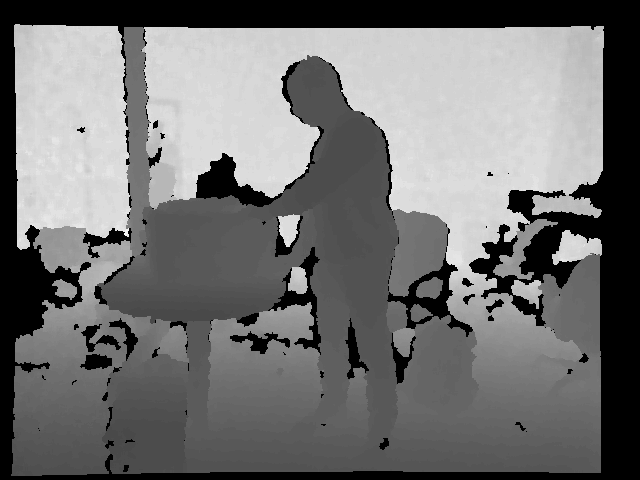}} \hspace{5mm}
\subfigure{\includegraphics[height=10mm]{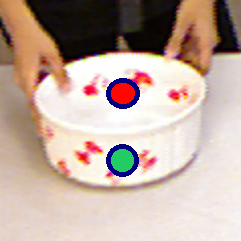}}
\subfigure{\includegraphics[height=10mm]{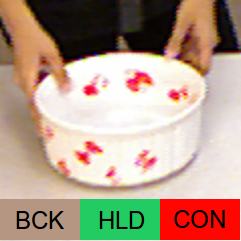}}

\caption{\label{fig:dataset_egs} (left) RGB-D image pairs illustrating images from (top row) the UMD turntable affordance dataset and (bottom row) the CAD120 dataset. (right) Illustrating the various levels of annotation (clockwise) original image, pixel level annotation, image level annotation, keypoint level annotation.}
\end{figure}

\section{Affordance Datasets}\label{sec:datasets}

There are not many datasets with pixelwise affordance labels. 
The RGB-D dataset proposed by~\cite{myers2015affordance} is an exception and focuses on part affordances of everyday tools.
The dataset consisting of 28,074 images is collected using a Kinect sensor, which records RGB and depth images at a resolution of $640\times480$ pixels and provides 7-class pixelwise affordance labels for objects from 17 categories. 
Each object is recorded on a revolving turntable to cover a full $360^\circ$ range of views providing clutter-free images of the object 
as shown in Figure~\ref{fig:dataset_egs}. 
While such lab recordings provide images with high quality, they lack important contextual information such as human-interaction.        

\begin{figure}[t]
\centering
\subfigure{\includegraphics[height=14mm]{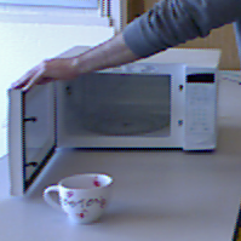}}
\subfigure{\includegraphics[height=14mm]{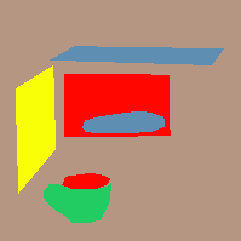}}
\subfigure{\includegraphics[height=14mm]{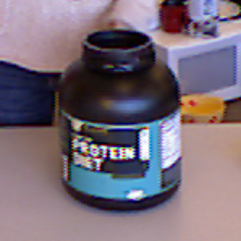}}
\subfigure{\includegraphics[height=14mm]{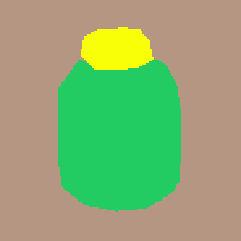}}
\subfigure{\includegraphics[height=14mm]{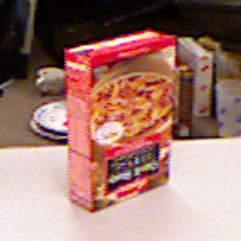}}
\subfigure{\includegraphics[height=14mm]{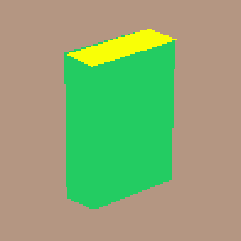}}
\subfigure{\includegraphics[height=14mm]{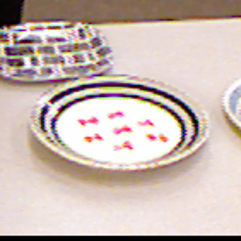}}
\subfigure{\includegraphics[height=14mm]{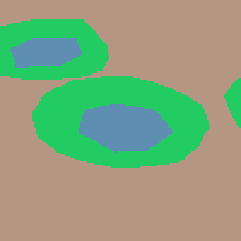}}
\subfigure{\includegraphics[height=14mm]{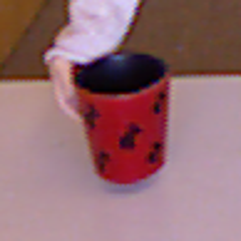}}
\subfigure{\includegraphics[height=14mm]{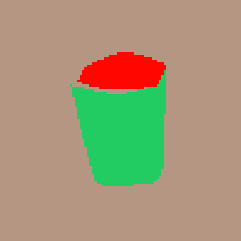}}
\subfigure{\includegraphics[height=14mm]{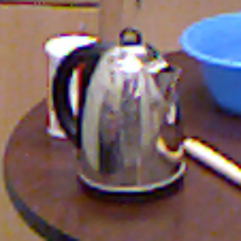}}
\subfigure{\includegraphics[height=14mm]{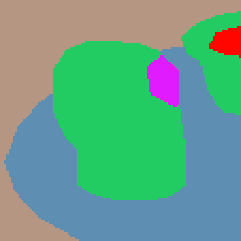}}
\subfigure{\includegraphics[height=14mm]{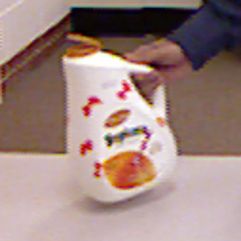}}
\subfigure{\includegraphics[height=14mm]{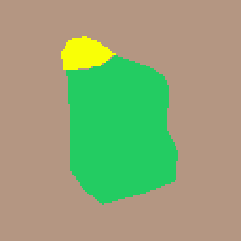}}
\subfigure{\includegraphics[height=14mm]{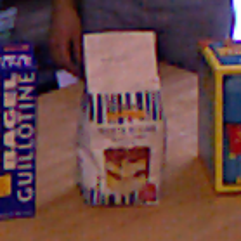}}
\subfigure{\includegraphics[height=14mm]{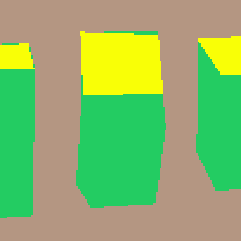}}
\subfigure{\includegraphics[height=14mm]{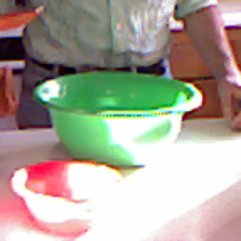}}
\subfigure{\includegraphics[height=14mm]{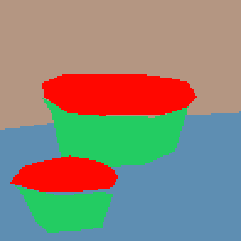}}
\subfigure{\includegraphics[height=14mm]{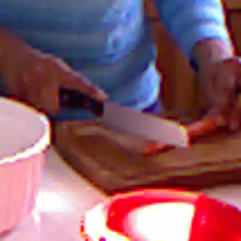}}
\subfigure{\includegraphics[height=14mm]{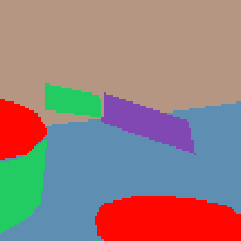}}
\subfigure{\includegraphics[height=14mm]{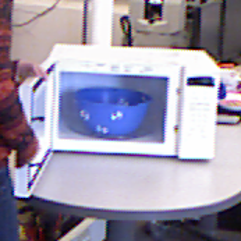}}
\subfigure{\includegraphics[height=14mm]{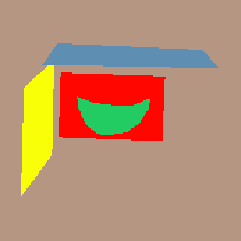}}
\subfigure{\includegraphics[height=14mm]{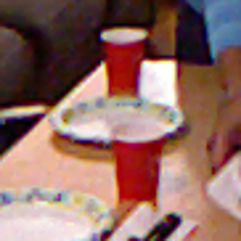}}
\subfigure{\includegraphics[height=14mm]{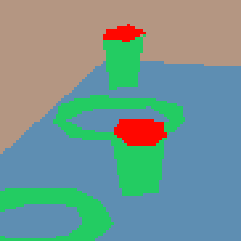}}
\caption{\label{fig:cad_120_annotation_examples} Sample images from the proposed CAD120 affordance dataset. The affordance labels are background (brown), holdable (green), openable (yellow), supportable (blue), containable (red), cuttable (purple) and pourable (magenta).}
\end{figure}

We therefore adopt a dataset that contains objects within the context of human-interactions in a more realistic environment. 
We found the CAD120 dataset~\cite{koppula2014physically} to be well tailored for our purpose. 
It consists of 215 videos in which 8 actors perform 14 different high-level activities. 
Each high-level activity is composed of sub-activities, which in turn involve one or more objects. 
In total, there are 32 different sub-activities and 35 object classes.
A few images of the dataset are shown in Figure~\ref{fig:dataset_egs}.
The dataset also provides framewise annotation of the sub-activity, object bounding boxes and automatically extracted human pose. 

We annotate the affordance labels \emph{openable, cuttable, containable, pourable, supportable, holdable} for every 10$^{th}$ frame from sequences involving an active human-object interaction resulting in 3090 frames. 
Each frame contains between 1 and 12 object instances resulting in 9916 objects in total. 
We annotate all object instances with pixelwise affordance labels. 
A few images from the dataset are shown in Figure~\ref{fig:cad_120_annotation_examples}.
As can be seen, the appearance of affordances can vary significantly \eg visually distinct object parts like the lid of a box, the cap of a bottle and the door of a microwave all have the affordance \textit{openable}. 
Similarly, the interiors of a bowl and a microwave are \textit{containable}.   

\begin{figure}[t]
\centering
\subfigure{\includegraphics[height=35mm]{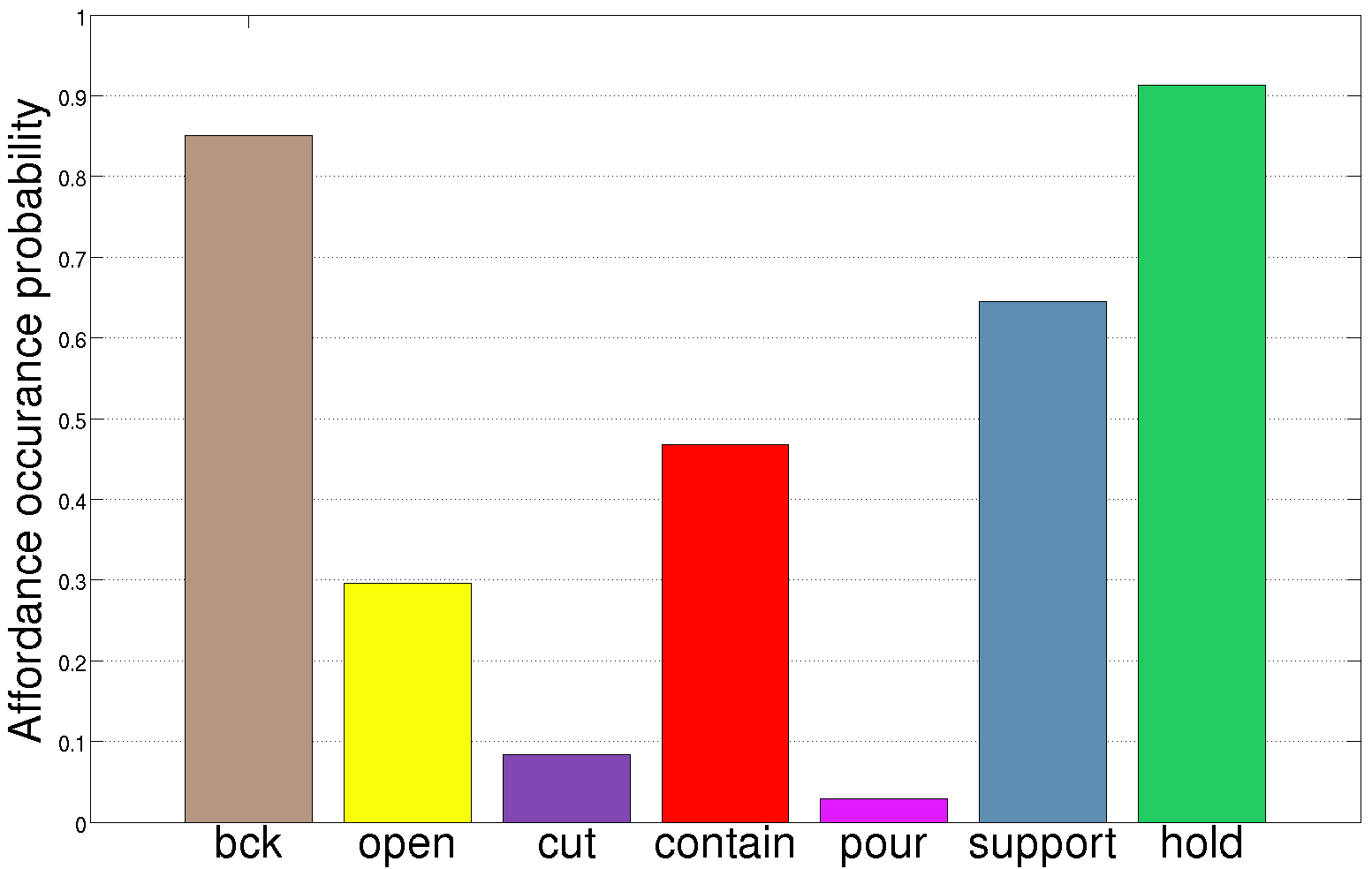}}
\subfigure{\includegraphics[height=35mm]{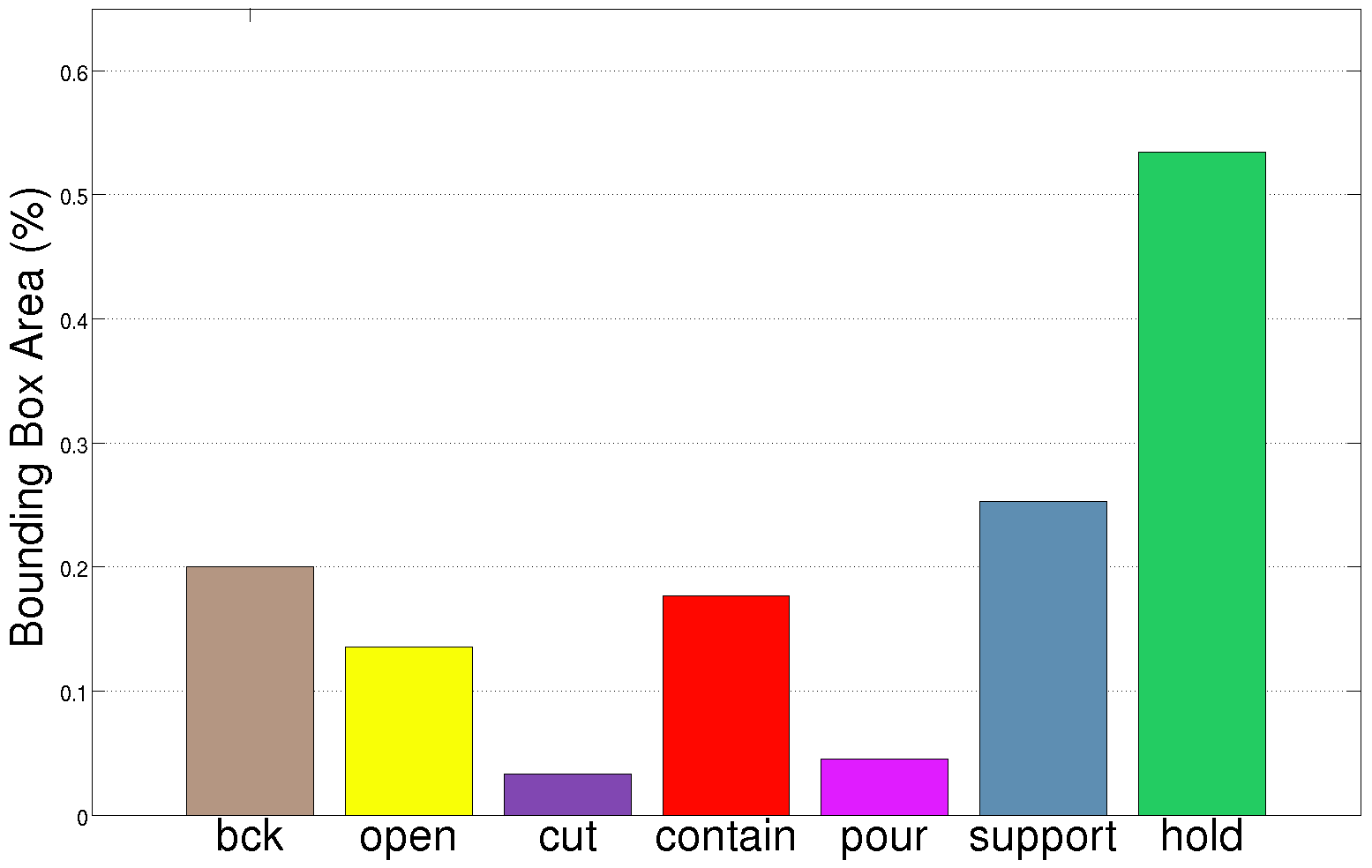}}
\caption{\label{fig:cad120_annotation_metadata} Distribution of affordance labels at the object bounding box level in the proposed dataset (left) probability of observing an affordance (right) median area covered by an affordance segment in relation to its object bounding box.}
\end{figure}

Figure~\ref{fig:cad120_annotation_metadata} presents statistics of affordance segments at the level of object bounding boxes. 
As can be seen, affordances \textit{holdable, supportable} are most likely to occur because most interacted objects are handheld in the context of a supportive structure. 
Also, affordances like \textit{openable, containable} which are a result of generic interactions have a fair chance to be observed. 
However, precise affordances like \textit{cuttable, pourable} not only occur rarely, but also cover a minuscule portion of the object bounding box. 
All other affordances are well represented visually in that they cover at least $15\%$ of the object bounding boxes, which have a median dimension of $68\times57$.
The dataset is also well balanced in terms of the number of images contributed by each actor with a median of 382 and a range of 227--606 images.
We intend to make this dataset publicly available. 

\section{Proposed Method}\label{sec:proposed_method}
Supervised learning of affordances using appearance features has been addressed in~\cite{katz2014perceiving,kim2014semantic,myers2015affordance,hermans2011affordance,lenz2015deep,song2015learning}. 
Recently, a supervised framework for object affordance labeling in RGB-D images is proposed by~\cite{myers2015affordance}.
The framework treats each class independently by learning standalone one-vs-all classifiers, affecting model scalability adversely. 
In this regard, owing to recent advances in end-to-end techniques for semantic image segmentation, we build on the DeepLab model~\cite{chen2014semantic}.
This method uses a DCNN to predict the label distribution per pixel, followed by a fully connected CRF to smooth predictions while preserving image edges. 

We now describe the learning procedure at various levels of supervision. 
Given an image $I$ with $n$ pixels, we denote the image values as $X = \{x_1,x_2,\dots,x_n\}$ and the corresponding labeling as $Y = \{y_1,y_2,\dots,y_n\}$ where $y_i \in \mathcal{L}$ takes one of the $L$ discrete labels $\mathcal{L} = \{1,2,\dots,L\}$ with $L=1$ indicating the background class. 
Note that these pixel level labels may not be available for the training set. 
Instead, we consider two cases of weak annotations. In the first case, a set of image level labels are provided. They are denoted by $Z = \{z_1,z_2,\dots\}$, where $z_l \in \mathcal{L}$ and $\sum_{i}[y_i=z_l] > 0$, \ie $Z$ contains the classes that are present anywhere in the image. In the second case, an additional reference point in the image is provided for each $z \in Z$. We denote this by $Z_x = \{(z_1,x_1),(z_2,x_2),\dots\}$ where $x_l$ is the pixel with label $z_l$.   
The latter case of weak annotation is based on single keypoints annotated by users. 
This technique has been used to scale up the annotation process for a large-scale material database in~\cite{bell2015material}. 
Figure~\ref{fig:dataset_egs} illustrates the various levels of annotation.

We will briefly summarize the supervised learning based on~\cite{chen2014semantic} in Section~\ref{sec:pixel}. 
In Section~\ref{sec:weak}, we propose an approach for weakly supervised learning and discuss its initialization in Section~\ref{sec:keypoints_annotation}. 
Finally, we propose an approach that transfers annotations of the type $Z_x$ to images with weaker annotations of type $Z$ in Section~\ref{sec:hpose_keypoint_regression}. 
We will exploit automatically extracted human pose as context for the annotation transfer. 

\subsection{Pixel level annotation}
\label{sec:pixel}
In the fully supervised case, the objective function is the log likelihood given by
\begin{equation} \label{eq:supervised_dcnn_loss_function}
 J(\theta) = \log P(Y|X;\theta) = \sum_{i=1}^{n} \log P(y_i|X;\theta),
\end{equation}
where $\theta$ is the vector of DCNN parameters. The per-pixel label distribution is then given by
\begin{equation} \label{eq:supervised_dcnn_pixel_label}
 P(y_i|X;\theta) \propto exp\left(f_i\left(y_i|X;\theta\right)\right),
\end{equation}
where $f_i\left(y_i|X;\theta\right)$ is the output of the DCNN at pixel $i$. 
For optimizing $J(\theta)$, we adopt the implementation provided by~\cite{chen2014semantic}.

\subsection{Weak annotation}
\label{sec:weak}
Considering the case when only weak image level annotation is available, the observed variables are image data $X$ and image level labels $Z$. 
The second case $Z_x$ is very similar and will be discussed in Section~\ref{sec:keypoints_annotation}. 
The pixel level segmentation $Y$ forms the latent variables. 
Our approach is based on the EM framework that has been proposed in~\cite{papandreou2015weakly}. 
While \cite{papandreou2015weakly} introduces class dependent bias terms that are constant for an entire image, \ie independent of the image location, we extend the framework to handle spatial dependencies. 
In this way, we are not limited to image level labels $Z$, but we can also use weak annotations of the second type $Z_x$.
 
We formulate an EM approach in order to learn the parameters $\theta$ of the DCNN model, which is given by
\begin{equation}
\begin{split}
 P(X,Y,Z;\theta)&=P(Y|X,Z;\theta) \> P(X,Z) \\
 &=\prod_{i=1}^{n}P(y_i|X,Z;\theta) \> P(X,Z).
\end{split}
\end{equation}
 
The M-step involves updating the model parameters $\theta$ by treating the point estimate $\hat{Y}$ as groundtruth segmentation and optimizing
\begin{equation}\label{eq:mstep}
\sum_{Y}P(Y|X,Z;\theta^{old})\;\log P(Y|Z;\theta) \approx \log P(\hat{Y}|X;\theta) = \sum_{i=1}^{n} \log P(\hat{y}_i|X;\theta),
\end{equation}
which can be efficiently performed by minibatch stochastic gradient descent (SGD) as in~\eqref{eq:supervised_dcnn_loss_function}.  

While the M-step is the same as in~\cite{papandreou2015weakly}, the E-step differs since we model spatial dependencies of the label distribution given $Z$. 
The E-step amounts to computing the point estimate $\hat{Y}$ of the latent segmentation as
\begin{equation}\label{eq:estep_approximation}
\begin{split}
\hat{Y} = \underset{Y}{\text{argmax}} \; \log P(Y|X,Z;\theta) &= \underset{\{y_1,\dots,y_n\}}{\text{argmax}} \; \sum_{i=1}^{n} \log P(y_i|X,Z;\theta) \\
\end{split}
\end{equation}
\begin{equation} \label{eq:affordance_gmm}
P(y_i|X,Z;\theta) = \begin{cases}
  \frac{f_{bg}}{f_{bg} + \sum_{z\in Z\setminus\{1\}} \sum_k \bm{\pi}_{z,k} \mathcal{N}(i;\bm{\mu}_{z,k}(X;\theta),\bm{\Sigma}_{z,k}(X;\theta))}, & \text{if}\ y_i = 1 \vspace{3mm} \\ 
  \frac{\sum_k \bm{\pi}_{y_i,k} \mathcal{N}(i;\bm{\mu}_{y_i,k}(X;\theta),\bm{\Sigma}_{y_i,k}(X;\theta))}{f_{bg} + \sum_{z\in Z\setminus\{1\}} \sum_k \bm{\pi}_{z,k} \mathcal{N}(i;\bm{\mu}_{z,k}(X;\theta),\bm{\Sigma}_{z,k}(X;\theta))}, & \text{if}\ y_i \in Z\setminus\{1\} \vspace{3mm} \\ 
  0, & \text{otherwise}
\end{cases} 
\end{equation}
where $\setminus$ indicates set subtraction. For the background class, we assume a spatially constant probability $f_{bg}$. For affordances that are not part of the weak labels $Z$ the probability is set to zero. 
The spatial distribution of an affordance $z \in Z$ is modeled by a Gaussian Mixture distribution with weights $\bm{\pi}_{z,k}$, means $\bm{\mu}_{z,k}$ and covariance matrices $\bm{\Sigma}_{z,k}$, which depend on $\theta$ and $X$. 
Given the output of the DCNN, \ie $P(y_i|X;\theta)$ from \eqref{eq:supervised_dcnn_pixel_label}, we compute the set of pixels that are labeled by $z$, \ie $\{ i : z= \text{argmax}_{y_i} P(y_i|X;\theta)\}$. 
A binary Grabcut segmentation is initialized with this set as foreground and the rest of the pixels as background. 
8-neighbor connected regions of size larger than $10\%$ of the largest region are considered to estimate parameters $\bm{\pi}_{z,k}$, $\bm{\mu}_{z,k}$ and $\bm{\Sigma}_{z,k}$.

\subsection{Initialization}
\label{sec:keypoints_annotation}

We consider two sets of training images. 
While the first set is annotated by a set of keypoints $Z_x$, a second set contains only image level labels $Z$ as shown in Figure~\ref{fig:dataset_egs}. 
We start with the first set annotated with $Z_x$ and initialize the Gaussian Mixture for $z_l$ by a single Gaussian with  $\bm{\mu}_{z_l} = x_l$ and $\bm{\Sigma}_{z_l} = 40 \bm{I}$.
We perform the E-step to learn the initial point estimate $\hat{Y}$ using \eqref{eq:estep_approximation}. 
The DCNN model is initialized by a pre-trained model VGG16~\cite{simonyan2014very} and we update the model parameters according to the M-step \eqref{eq:mstep}. 
The updated DCNN model is then applied to all training images to compute \eqref{eq:supervised_dcnn_pixel_label} and we continue with the E-step. 
For the keypoints $(z_l,x_l)$, we retain the means of the Gaussians $\bm{\mu}_{z_l}$ as $x_l$.
The approach is then iterated until convergence. 

\subsection{Estimating keypoints from human pose} 
\label{sec:hpose_keypoint_regression}

Since affordances can be observed in the context of human-object interaction, we propose to transfer keypoint annotations to the set that contains only image level labels. 
Given the automatically extracted 2d human pose and detected bounding boxes of objects, we represent the human pose $h$ as a $2J$ dimensional vector of joint locations where $J$ denotes the number of joints. 
For each affordance $z \in \mathcal{L}$, we collect all annotated keypoints $x_t$ together with the pose $h_t$ as a training set. 
We further normalize the pose vector $h_t$ and $x_t$ by subtracting the center of the object bounding box followed by mean and variance normalization over the training data, \ie setting mean to zero and standard deviation to one.
We then perform k-means clustering on these poses to learn a dictionary of size $D$, denoted as $h_D$.

For regressing the normalized keypoint $x$ of an affordance $z$ given the normalized pose $h$, we use a regularized non-linear regression with an RBF kernel
\begin{equation}\label{eq:hpose_to_keypoint_regression}
x = \bm{\alpha}^T \bm{\phi}(h,h_D) = \sum_{d=1}^{D} \alpha_d \exp\left(-\frac{\|h-h_d\|_2^2}{\gamma^2}\right) 
\end{equation}
where $h_d$ is the $d^{th}$ entry of dictionary $h_D$. 
The regression weight $\bm{\alpha}$ is learned in the least squared error sense. 
Hyperparameter $\gamma$, regularization parameter $\lambda$ and dictionary size $D$ are learned through cross validation. 

\section{Experiments}\label{sec:experiments}
We first evaluate the fully supervised approach (Section~\ref{sec:pixel}) and compare it with other fully supervised approaches for affordance segmentation. We then compare the discussed weakly supervised settings (Section~\ref{sec:weak}) with the fully supervised baseline.
We evaluate on two affordance datasets presented in Section~\ref{sec:datasets}. 
As evaluation protocol, we follow the predefined train-test split for the UMD turntable dataset.
Regarding the CAD120 affordance dataset, we reserve images from actors $\{5,9\}$ as $test$ and refer to images from actors $\{\{1,6\},\{3,7\},\{4,8\}\}$ as $\{TrainA, TrainB, TrainC\}$ respectively. 
Further, we refer to the union of the three training sets as \textit{allTrain}.

For quantitative evaluation, we report per class intersection-over-union (IoU), which is also known as Jaccard index, for both datasets. 
Since~\cite{myers2015affordance} reports performance in terms of a Weighted F-Measure, we also report this metric for the UMD turntable dataset. 

\subsection{UMD Turntable Dataset}
In~\cite{myers2015affordance}, two approaches have been presented for learning affordances from local appearance and geometric features.
The first approach is based on features derived from a superpixel based hierarchical matching pursuit (HMP) together with a linear SVM
and the second approach is based on curvature and normal features derived from depth data used within a structured random forest (SRF).
We trained the DeepLab~\cite{chen2014semantic} framework initialied by VGG16~\cite{simonyan2014very} in the fully supervised setting. 
For SGD, we use a mini-batch of 6 images and an initial learning rate of 0.001 (0.01 for the final classifier layer), multiplying the learning rate by 0.1 after every 2000 iterations. 
We use a momentum of 0.9, weight decay of 0.0005 and run for 6000 iterations. 
The performance comparison on both IoU and weighted F-measure metrics are shown in Table~\ref{tab:umd_dataset_results}.

\begin{figure}[b!]
\centering
\subfigure{\includegraphics[height=14mm]{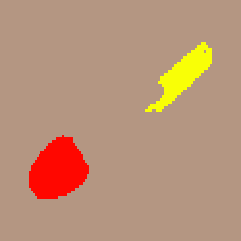}}
\subfigure{\includegraphics[height=14mm]{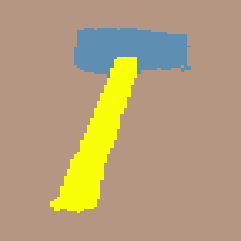}}
\subfigure{\includegraphics[height=14mm]{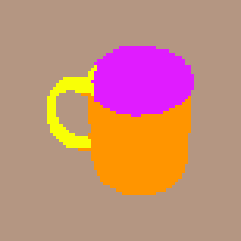}}
\subfigure{\includegraphics[height=14mm]{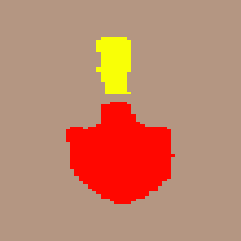}}
\subfigure{\includegraphics[height=14mm]{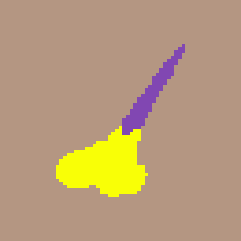}}
\subfigure{\includegraphics[height=14mm]{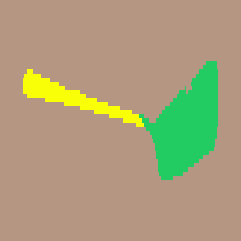}}
\subfigure{\includegraphics[height=14mm]{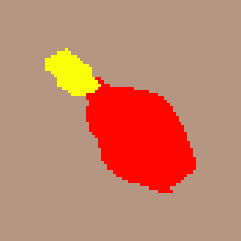}}
\subfigure{\includegraphics[height=14mm]{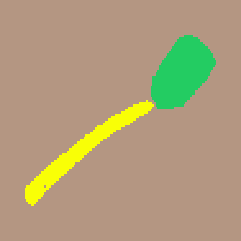}}
\subfigure{\includegraphics[height=14mm]{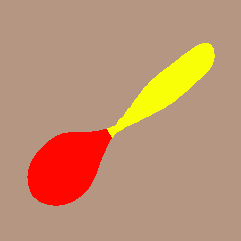}}
\subfigure{\includegraphics[height=14mm]{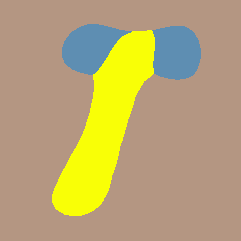}}
\subfigure{\includegraphics[height=14mm]{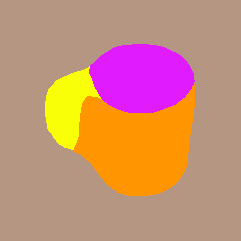}}
\subfigure{\includegraphics[height=14mm]{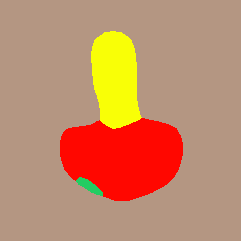}}
\subfigure{\includegraphics[height=14mm]{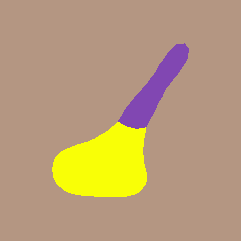}}
\subfigure{\includegraphics[height=14mm]{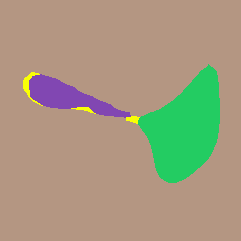}}
\subfigure{\includegraphics[height=14mm]{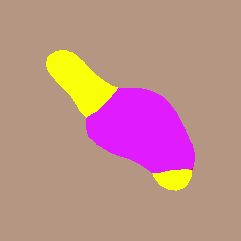}}
\subfigure{\includegraphics[height=14mm]{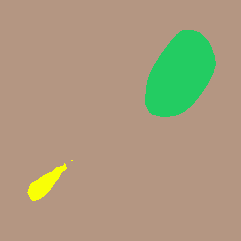}}
\subfigure{\includegraphics[height=14mm]{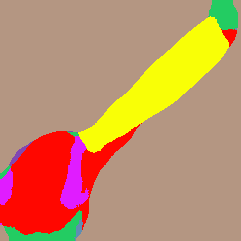}}
\subfigure{\includegraphics[height=14mm]{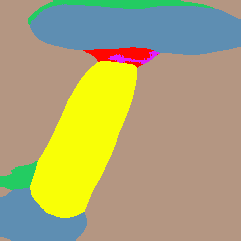}}
\subfigure{\includegraphics[height=14mm]{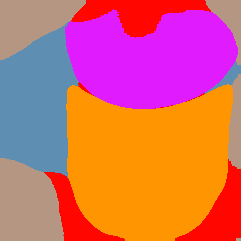}}
\subfigure{\includegraphics[height=14mm]{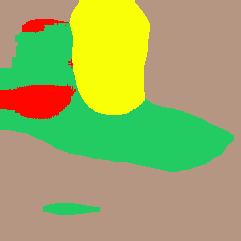}}
\subfigure{\includegraphics[height=14mm]{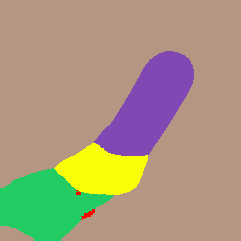}}
\subfigure{\includegraphics[height=14mm]{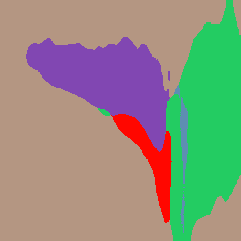}}
\subfigure{\includegraphics[height=14mm]{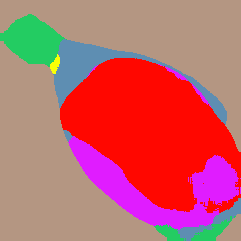}}
\subfigure{\includegraphics[height=14mm]{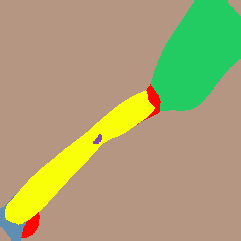}}
\caption{\label{fig:umd_qualitative_results} Results from supervised learning on UMD turntable dataset using (top) groundtruth segmentation (middle) DeepLab (bottom) DEP+SRF.}
\end{figure}

As can be observed, the trend in performance is similar irrespective of the evaluation metric. 
The \texttt{HMP+SVM} consistently outperforms the \texttt{DEP+SRF} combination, indicating that learning features from data is more effective than learning complex classifiers on handcrafted features. 
DeepLab in turn outperforms \texttt{HMP+SVM} in almost all classes reconfirming the effectiveness of end-to-end learning. 
However, in spite of powerful learning techniques, the performance for small affordance regions like $pound, support$ is considerably low.
A few qualitative results are shown in Figure~\ref{fig:umd_qualitative_results}.

\begin{table}[t]
\begin{center}
\scriptsize
\begin{tabular}{|c|C{12mm}|C{12mm}|C{12mm}|C{12mm}|C{12mm}|C{12mm}|C{12mm}|}
\hline
UMD Turntable & Grasp & Cut & Scoop & Contain & Pound & Support & Wgrasp \\
\hline\hline
 \multicolumn{8}{|>{\columncolor[gray]{.8}}c|}{Weighted F-Measure} \\
\hline\hline
 HMP + SVM & 0.37 & 0.37 & 0.42 & 0.81 & \textbf{0.64} & 0.52 & 0.77 \\ \hline
 DEP + SRF & 0.31 & 0.28 & 0.41 & 0.63 & 0.43 & 0.48 & 0.66 \\ \hline
 DeepLab   & \textbf{0.59} & \textbf{0.71} & \textbf{0.55} & \textbf{0.90} & 0.33 & \textbf{0.70} & \textbf{0.87} \\ \hline
\hline
 \multicolumn{8}{|>{\columncolor[gray]{.8}}c|}{IoU} \\
\hline\hline
 HMP + SVM & 0.31 & 0.11 & 0.07 & 0.30 & 0.06 & 0.06 & 0.17 \\ \hline
 DEP + SRF & 0.26 & 0.01 & 0.06 & 0.28 & 0.04 & 0.03 & 0.19 \\ \hline
 DeepLab   & \textbf{0.51} & \textbf{0.63} & \textbf{0.49} & \textbf{0.85} & \textbf{0.26} & \textbf{0.63} & \textbf{0.80} \\ \hline
\end{tabular}
\caption{\label{tab:umd_dataset_results} Evaluating fully supervised approaches for affordance segmentation on the UMD turntable dataset. Evaluation metrics based on weigted F-measure and IoU. HMP+SVM and DEP+SRF are proposed in~\cite{myers2015affordance} and DeepLab in~\cite{chen2014semantic}.}
\end{center}
\end{table}

\subsection{CAD120 Affordance Dataset}
Since object bounding boxes in the dataset are pre-annotated, we perform all experiments on cropped images after extending the bounding boxes by 30px (or maximum possible) in each direction. 

\subsubsection{Supervised training:} We first evaluate and compare the fully supervised approach. 
In this setting, we use the entire training data \emph{allTrain} with pixelwise label annotations as training set.   
The approaches proposed in~\cite{myers2015affordance} are based on depth data. 
However, owing to noisy depth images from the dataset, we found the performance of both approaches with depth features to be substantially lower than that with CNN features.
We therefore report the accuracy for the best setting, which is obtained by using features from all layers~\cite{hariharan2015hypercolumns} of the VGG16~\cite{simonyan2014very} network for the SRF. 
It must be noted that DeepLab~\cite{papandreou2015weakly} is also finetuned from a similar network. 
Referring to Table~\ref{tab:cad120_dataset_results}, we can see that DeepLab performs at a mean IoU of 0.42 whereas the SRF based approach performs at 0.26.
In spite of the high quality of annotation and the rich model, the accuracy for the classes $cut$ and $pour$ is almost zero for all methods. 
This is because of the small spatial extent of both classes 
and the lack of training data (see Figure~\ref{fig:cad120_annotation_metadata}). 
Qualitative results for both methods are presented in Figure~\ref{fig:cad120_qualitative_results}.

\subsubsection{Weakly supervised training with keypoint annotations:} In the second setting, we replace pixel-wise label annotations by keypoint annotations \ie we use the entire \emph{allTrain} with $Z_x$ annotations as the training set.
For the first experiment, we initialize our approach as described in Section \ref{sec:keypoints_annotation} where we initialize the Gaussians based on the keypoints and perform a single E- and M-step each. 
This is denoted as \texttt{allTrain+EM(1 iter.)} in Table~\ref{tab:cad120_dataset_results}. 
Compared to the fully supervised setting, the mean accuracy decreases from $0.42$ to $0.28$. 
We found the proposed EM approach to converge within 3--4 iterations resulting in increased accuracy for all classes.
This is denoted as \texttt{allTrain+EM}.
The largest improvement can be observed for the class \emph{support}, which increases from $0.35$ to $0.44$. 
The mean accuracy increases from $0.28$ to $0.31$.

In \eqref{eq:estep_approximation}, we model the spatial distributions of the affordances by Gaussian mixture distributions. 
As a heuristic, we could also use the output of the Grabcut segmentation as $\hat{Y}$. 
This approach is denoted as \texttt{allTrain+EM(1 iter.)+onlyGC}. 
Similarly, we can skip the Grabcut segmentation and estimate Gaussian mixture parameters directly from~\eqref{eq:supervised_dcnn_pixel_label}, denoted as \texttt{allTrain+EM(1 iter.)+onlyGM}.
The substantial drop in performance in both cases indicates that these components are critical for performance. 

\begin{table}[b!]
\begin{center}
\scriptsize
\begin{tabular}{|c||C{12mm}|C{12mm}|C{12mm}|C{12mm}|C{12mm}||C{12mm}|}
\hline
Experiment Setting & Bck & Open & Contain & Support & Hold & Mean \\
\hline \hline
 \multicolumn{7}{|>{\columncolor[gray]{.8}}c|}{Supervised Training on allTrain} \\ 
\hline \hline
 DeepLab~\cite{chen2014semantic}   & 0.75 & 0.46 & 0.52 & 0.64 & 0.60 & 0.42  \\ \hline
 VGG + SRF~\cite{myers2015affordance} & 0.62 & 0.20 & 0.22 & 0.39 & 0.39 & 0.26 \\ \hline
\hline
\multicolumn{7}{|>{\columncolor[gray]{.8}}c|}{Weakly Supervised Training on allTrain with Keypoints} \\
\hline \hline
 allTrain EM(1 iter.) & 0.65 & 0.27 & 0.30 & 0.35 & 0.38 & 0.28 \\ \hline
 allTrain EM & 0.67 & 0.29 & 0.34 & 0.44 & 0.42 & 0.31 \\ \hline
 allTrain EM(1 iter.)+onlyGC & 0.60 & 0.19 & 0.19 & 0.23 & 0.30 & 0.22 \\ \hline
 allTrain EM(1 iter.)+onlyGM & 0.40 & 0.19 & 0.09 & 0.27 & 0.20 & 0.16 \\
 \hline \hline
 \multicolumn{7}{|>{\columncolor[gray]{.8}}c|}{Weakly Supervised Training on TrainX with Keypoints} \\ \hline \hline
 TrainXOnly+EM(1 iter.) & 0.48 & 0.17 & 0.20 & 0.41 & 0.31 & 0.22 \\ \hline
  TrainXOnly+EM & 0.58 & 0.20 & 0.26 & 0.33 & 0.40 & 0.25 \\ \hline\hline
 \multicolumn{7}{|>{\columncolor[gray]{.8}}c|}{Weakly Supervised Training on TrainX with Keypoints and rest with Image Labels} \\ \hline \hline
 Semi+TrainX+DeepLab \cite{papandreou2015weakly}   & 0.42 & 0.17 & 0.09 & 0.26 & 0.20 & 0.16  \\ \hline 
 TrainX+Pose+EM(1 iter.)    & 0.61 & 0.22 & 0.24 & 0.33 & 0.37 & 0.25  \\ \hline
 TrainX+Pose+EM & 0.63 & 0.21 & 0.24 & 0.39 & 0.39 & 0.27  \\ \hline
 TrainX+EM & 0.38 & 0.17 & 0.21 & 0.39 & 0.29 & 0.21  \\ \hline
 TrainX+BB+EM(1 iter.) & 0.34 & 0.14 & 0.12 & 0.26 & 0.17 & 0.15 \\ \hline
\end{tabular}
\caption{\label{tab:cad120_dataset_results} Evaluating affordance segmentation on the CAD120 affordance dataset under various settings. The evaluation metric used is IoU. While the mean is computed over all classes, class results are shown only for a subset of classes.}
\end{center}
\end{table}

\subsubsection{Weakly supervised training with mixed keypoint and image annotations:} In the third setting, we have two sets of training data.
The first set is annotated by keypoints $Z_x$ and the second set is annotated by image labels $Z$. 
We perform an evaluation averaged over three splits. 
For a split, one of the subsets \textit{trainA}, \textit{trainB}, or \textit{trainC} is annotated with $Z_x$ and the other subsets are annotated with $Z$. 

To begin with, we train our approach only on \textit{trainX}, \ie the subset annotated with $Z_x$, and do not use the training images annotated with $Z$. The approach is denoted as \texttt{TrainXOnly+EM(1 iter.)} in Table~\ref{tab:cad120_dataset_results}. 
As expected, the reduction of training data by one third decreases the mean accuracy from $0.28$ to $0.22$. 
Running the proposed EM approach until convergence improves the results by $3\%$ to $0.25$, denoted as \texttt{TrainXOnly+EM}.
The approach serves as baseline for other weakly supervised approaches that use additional training data annotated by $Z$.  

For comparison, we use the approach~\cite{papandreou2015weakly}. 
We initialize the method on \textit{TrainX} in the same way as the proposed method and set the fg--bg bias to 0.3--0.2. 
We achieve the best result using the semi-supervised mode of~\cite{papandreou2015weakly} where the initial segmentation results on \textit{TrainX} are not changed. 
This performs with a mean accuracy of only $0.16$ (\texttt{Semi+TrainX+DeepLab}), which is lower than the baseline \texttt{TrainXonly+EM}. 
This shows that constant bias terms proposed for the E-step in~\cite{papandreou2015weakly} are insufficient for the task of affordance segmentation.    

We now study the effect of transferring keypoints $Z_x$ which are available for \textit{TrainX} to other images in \textit{allTrain} using the method described in Section~\ref{sec:hpose_keypoint_regression}. 
The parameters of the regression in \eqref{eq:hpose_to_keypoint_regression} were obtained by cross validation. 
This resulted in $D=200$, $\gamma=10$, $\lambda=1$ which are used for all experiments. 
Referring to \texttt{TrainX+Pose+EM(1 iter.)}, the keypoint transfer based on human context followed by a single EM iteration improves the mean accuracy by $3\%$ to $0.25$. 
The performance further increases to $0.27$ when iterated until convergence. 
The same experiment performed without keypoint regression \ie using \textit{TrainXOnly} for the first EM iteration and using all training images for further iterations resulted in a slight drop in performance to $0.21$ (\texttt{TrainX+EM}).

In order to demonstrate that the performance gain is indeed from using human pose, we repeat the above experiment but regress keypoints from object bounding boxes instead of human pose. 
To this end, we replace the $2J$ dimensional pose vector $h_t$ by a 6d vector of the bounding box consisting of the x- and y-coordinates of the top left corner, width and height of the object bounding box. 
This setting, tabulated as \texttt{TrainX+BB+EM}, performs substantially worse, showing that human pose provides a valuable source for weakly supervised learning of affordances.  

\begin{figure}[h!]
\centering
\subfigure{\includegraphics[height=16mm]{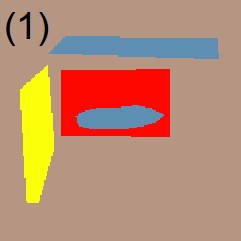}}
\subfigure{\includegraphics[height=16mm]{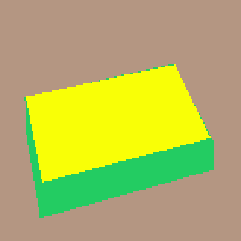}}
\subfigure{\includegraphics[height=16mm]{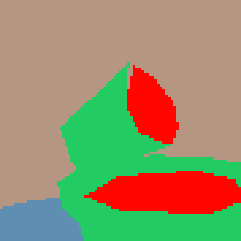}}
\subfigure{\includegraphics[height=16mm]{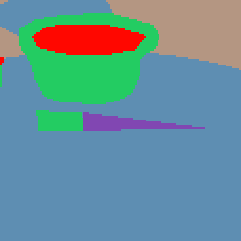}}
\subfigure{\includegraphics[height=16mm]{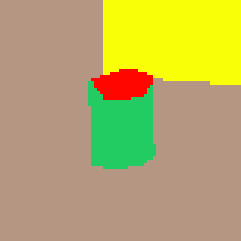}}
\subfigure{\includegraphics[height=16mm]{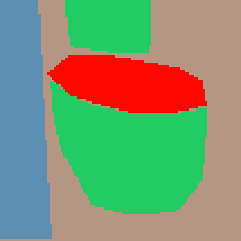}}
\subfigure{\includegraphics[height=16mm]{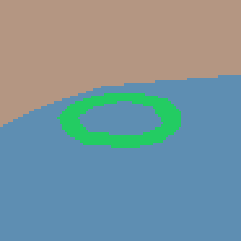}}

\subfigure{\includegraphics[height=16mm]{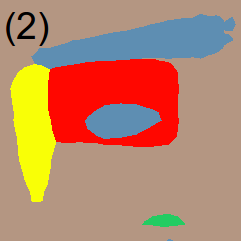}}
\subfigure{\includegraphics[height=16mm]{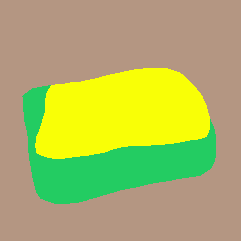}}
\subfigure{\includegraphics[height=16mm]{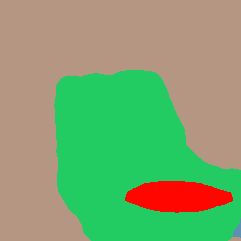}}
\subfigure{\includegraphics[height=16mm]{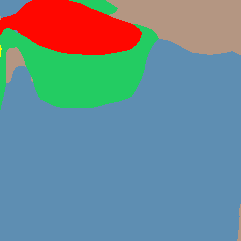}}
\subfigure{\includegraphics[height=16mm]{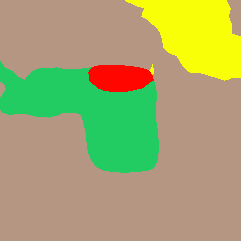}}
\subfigure{\includegraphics[height=16mm]{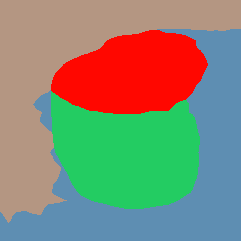}}
\subfigure{\includegraphics[height=16mm]{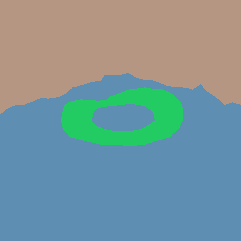}}

\subfigure{\includegraphics[height=16mm]{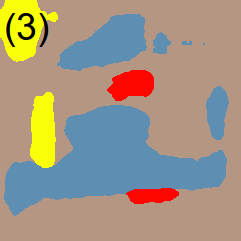}}
\subfigure{\includegraphics[height=16mm]{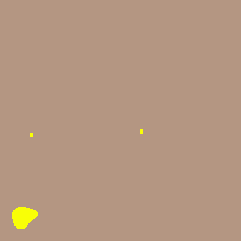}}
\subfigure{\includegraphics[height=16mm]{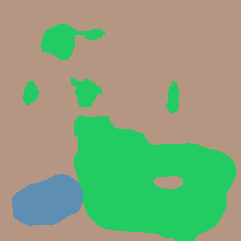}}
\subfigure{\includegraphics[height=16mm]{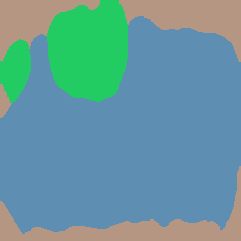}}
\subfigure{\includegraphics[height=16mm]{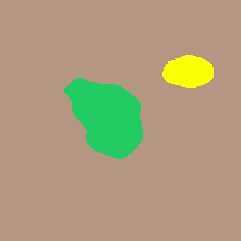}}
\subfigure{\includegraphics[height=16mm]{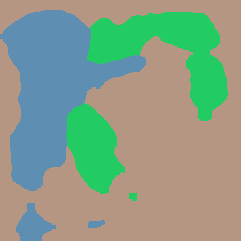}}
\subfigure{\includegraphics[height=16mm]{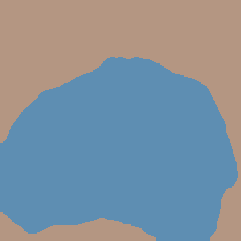}}

\subfigure{\includegraphics[height=16mm]{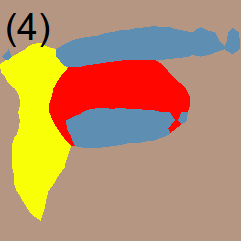}}
\subfigure{\includegraphics[height=16mm]{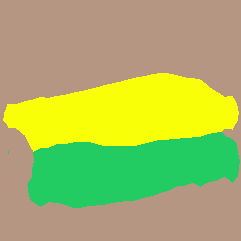}}
\subfigure{\includegraphics[height=16mm]{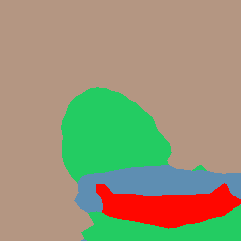}}
\subfigure{\includegraphics[height=16mm]{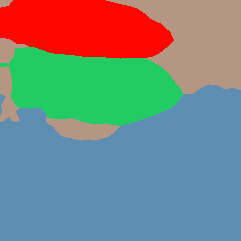}}
\subfigure{\includegraphics[height=16mm]{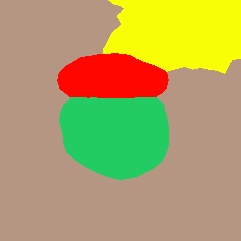}}
\subfigure{\includegraphics[height=16mm]{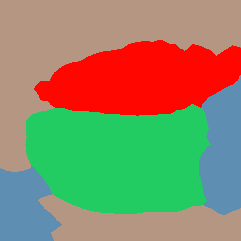}}
\subfigure{\includegraphics[height=16mm]{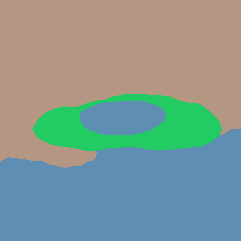}}

\subfigure{\includegraphics[height=16mm]{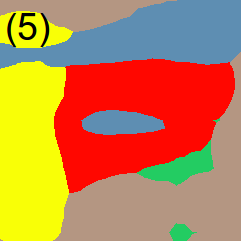}}
\subfigure{\includegraphics[height=16mm]{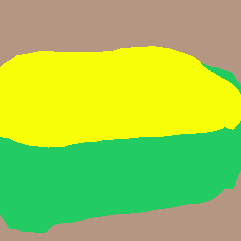}}
\subfigure{\includegraphics[height=16mm]{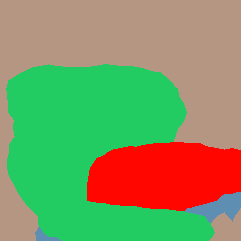}}
\subfigure{\includegraphics[height=16mm]{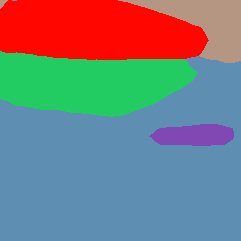}}
\subfigure{\includegraphics[height=16mm]{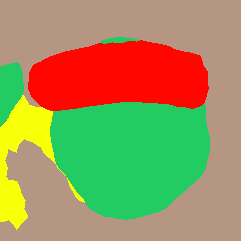}}
\subfigure{\includegraphics[height=16mm]{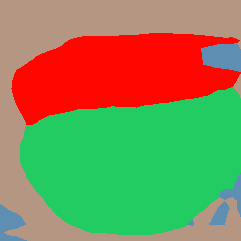}}
\subfigure{\includegraphics[height=16mm]{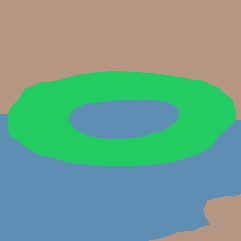}}

\subfigure{\includegraphics[height=16mm]{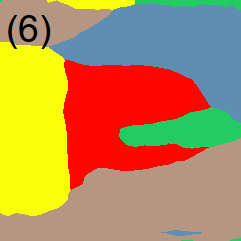}}
\subfigure{\includegraphics[height=16mm]{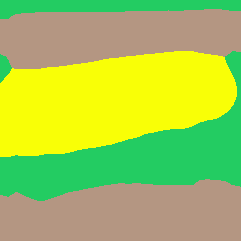}}
\subfigure{\includegraphics[height=16mm]{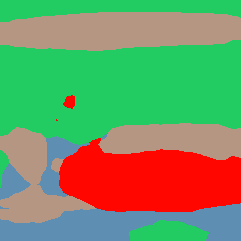}}
\subfigure{\includegraphics[height=16mm]{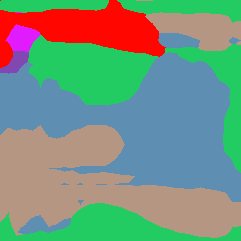}}
\subfigure{\includegraphics[height=16mm]{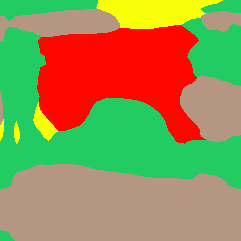}}
\subfigure{\includegraphics[height=16mm]{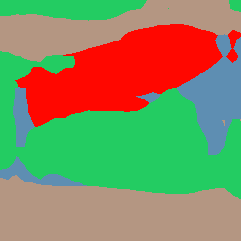}}
\subfigure{\includegraphics[height=16mm]{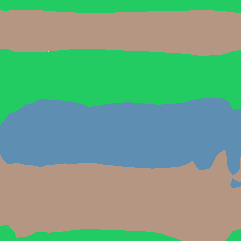}}

\subfigure{\includegraphics[height=16mm]{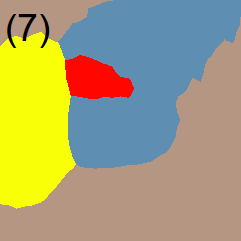}}
\subfigure{\includegraphics[height=16mm]{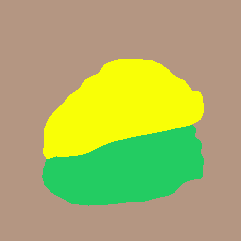}}
\subfigure{\includegraphics[height=16mm]{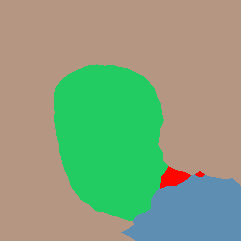}}
\subfigure{\includegraphics[height=16mm]{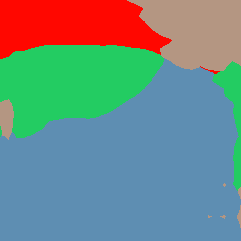}}
\subfigure{\includegraphics[height=16mm]{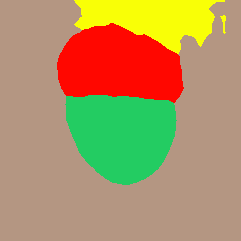}}
\subfigure{\includegraphics[height=16mm]{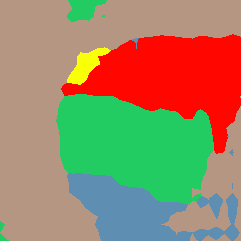}}
\subfigure{\includegraphics[height=16mm]{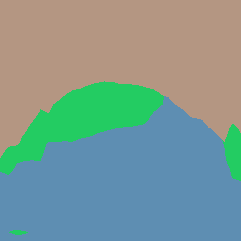}}

\subfigure{\includegraphics[height=16mm]{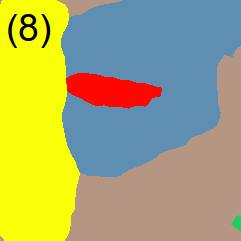}}
\subfigure{\includegraphics[height=16mm]{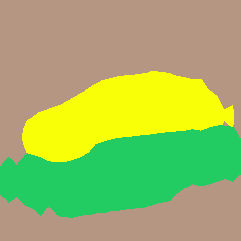}}
\subfigure{\includegraphics[height=16mm]{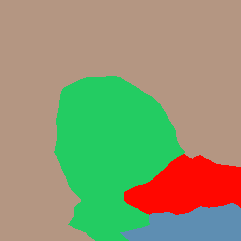}}
\subfigure{\includegraphics[height=16mm]{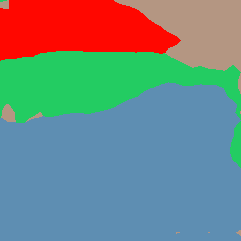}}
\subfigure{\includegraphics[height=16mm]{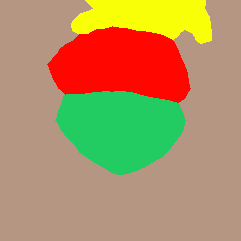}}
\subfigure{\includegraphics[height=16mm]{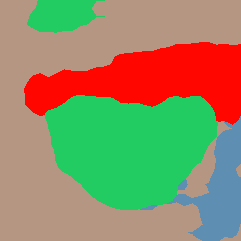}}
\subfigure{\includegraphics[height=16mm]{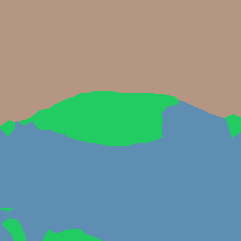}}

\subfigure{\includegraphics[height=16mm]{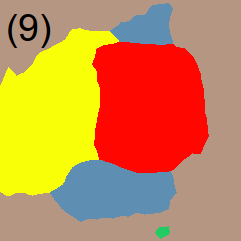}}
\subfigure{\includegraphics[height=16mm]{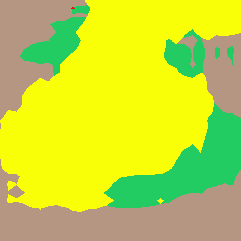}}
\subfigure{\includegraphics[height=16mm]{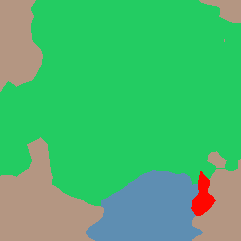}}
\subfigure{\includegraphics[height=16mm]{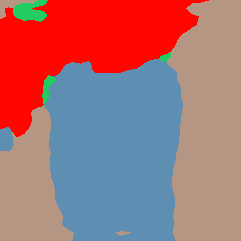}}
\subfigure{\includegraphics[height=16mm]{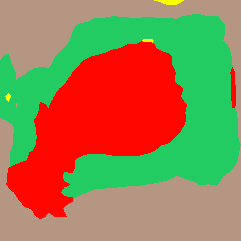}}
\subfigure{\includegraphics[height=16mm]{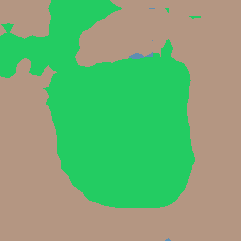}}
\subfigure{\includegraphics[height=16mm]{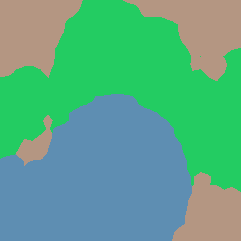}}

\caption{\label{fig:cad120_qualitative_results} Results from CAD120 affordance dataset as presented in Table~\ref{tab:cad120_dataset_results}. (1) Ground truth (2) Supervised \texttt{DeepLab}\cite{chen2014semantic} (3) Supervised \texttt{VGG+SRF}\cite{myers2015affordance} (4) \texttt{allTrain+EM(1 Iter.)} (5) \texttt{TrainX+EM(1 Iter.)} (6) \texttt{Semi+TrainX+DeepLab}\cite{papandreou2015weakly} (7) \texttt{TrainX+Pose+EM(1 Iter.)} (8) \texttt{TrainX+Pose+EM} (9) \texttt{TrainX+BB+EM(1 Iter.)} Image best viewed in color.}
\end{figure}

Figure~\ref{fig:cad120_qualitative_results} presents qualitative results of various discussed approaches presented in Table~\ref{tab:cad120_dataset_results}.
In the supervised setting, segmentations generated by \texttt{DeepLab} are greatly superior to those generated from \texttt{VGG+SRF}, which performs poorly even for affordances with large spatial extent like \textit{containable, openable}.
However, both approaches perform poorly for difficult affordances like \textit{cuttable}. 
Regarding weakly supervised setting with keypoints, there is a visible drop of quality for \texttt{allTrain+EM(1 Iter.)} when compared to the fully supervised approach as expected. 
A further degradation is seen with \texttt{TrainXOnly+EM(1 Iter.)} due to the reduced training data. 
Comparing weakly supervised approaches with mixed annotations, \texttt{Semi+TrainX+DeepLab}\cite{chen2014semantic} 
allocates equally sized segments for all affordances. 
By contrast, improvements due to the proposed EM approach 
can already be seen for \texttt{TrainXOnly+EM(1 Iter.)}. 
Further, improved spatial localization of affordances due to regressing keypoints from human pose is seen for \texttt{TrainX+HPose+EM(1 Iter.)} which is further refined by \texttt{TrainX+HPose+EM}. 
Finally, the last row shows the poor performance of \texttt{TrainX+BB+EM(1 Iter)}.
When compared with \texttt{TrainX+HPose+EM(1 Iter.)}, this indicates the importance of keypoint transfer based on human pose.

\section{Conclusion}
In this work, we have addressed the problem of weakly supervised affordance segmentation. To this end, we proposed an expectation-maximization approach that can be trained on weak keypoint annotations. In addition, we showed how contextual information from human-object interaction can be used to transfer such annotations to images with only image level annotations. This improved the segmentation accuracy of our EM approach substantially. For evaluation, we introduced a pixel-wise annotated affordance dataset containing 3090 images and 9916 object instances with rich contextual information which can be used to further investigate the impact of context on affordance segmentation.

\bibliographystyle{splncs}
\bibliography{egbib}

\begin{thebibliography}{10}

\bibitem{parikh2011relative}
Parikh, D., Grauman, K.:
\newblock Relative attributes.
\newblock In: ICCV. (2011)  503--510

\bibitem{liu2011recognizing}
Liu, J., Kuipers, B., Savarese, S.:
\newblock Recognizing human actions by attributes.
\newblock In: CVPR. (2011)  3337--3344

\bibitem{patterson2012sun}
Patterson, G., Hays, J.:
\newblock Sun attribute database: Discovering, annotating, and recognizing
  scene attributes.
\newblock In: CVPR. (2012)  2751--2758

\bibitem{chen2015deepdriving}
Chen, C., Seff, A., Kornhauser, A., Xiao, J.:
\newblock Deepdriving: Learning affordance for direct perception in autonomous
  driving.
\newblock In: ICCV. (2015)  2722--2730

\bibitem{koppula2013learning}
Koppula, H.S., Gupta, R., Saxena, A.:
\newblock Learning human activities and object affordances from rgb-d videos.
\newblock IJRR \textbf{32}(8) (2013)  951--970

\bibitem{katz2014perceiving}
Katz, D., Venkatraman, A., Kazemi, M., Bagnell, J.A., Stentz, A.:
\newblock Perceiving, learning, and exploiting object affordances for
  autonomous pile manipulation.
\newblock Autonomous Robots \textbf{37}(4) (2014)  369--382

\bibitem{kim2014semantic}
Kim, D.I., Sukhatme, G.:
\newblock Semantic labeling of 3d point clouds with object affordance for robot
  manipulation.
\newblock In: ICRA. (2014)  5578--5584

\bibitem{myers2015affordance}
Myers, A., Teo, C.L., Fermuller, C., Aloimonos, Y.:
\newblock Affordance detection of tool parts from geometric features.
\newblock In: ICRA. (2015)  1374--1381

\bibitem{hermans2011affordance}
Hermans, T., Rehg, J.M., Bobick, A.:
\newblock Affordance prediction via learned object attributes.
\newblock In: ICRA: Workshop on Semantic Perception, Mapping, and Exploration.
  (2011)

\bibitem{papandreou2015weakly}
Papandreou, G., Chen, L.C., Murphy, K.P., Yuille, A.L.:
\newblock Weakly-and semi-supervised learning of a deep convolutional network
  for semantic image segmentation.
\newblock In: ICCV. (2015)  1742--1750

\bibitem{bell2015material}
Bell, S., Upchurch, P., Snavely, N., Bala, K.:
\newblock Material recognition in the wild with the materials in context
  database.
\newblock In: CVPR. (2015)  3479--3487

\bibitem{khan2012color}
Khan, F.S., Anwer, R.M., van~de Weijer, J., Bagdanov, A.D., Vanrell, M., Lopez,
  A.M.:
\newblock Color attributes for object detection.
\newblock In: CVPR. (2012)  3306--3313

\bibitem{farhadi2009describing}
Farhadi, A., Endres, I., Hoiem, D., Forsyth, D.:
\newblock Describing objects by their attributes.
\newblock In: CVPR. (2009)  1778--1785

\bibitem{lampert2009learning}
Lampert, C.H., Nickisch, H., Harmeling, S.:
\newblock Learning to detect unseen object classes by between-class attribute
  transfer.
\newblock In: CVPR. (2009)  951--958

\bibitem{ferrari2007learning}
Ferrari, V., Zisserman, A.:
\newblock Learning visual attributes.
\newblock In: NIPS. (2007)  433--440

\bibitem{zhu2014reasoning}
Zhu, Y., Fathi, A., Fei-Fei, L.:
\newblock Reasoning about object affordances in a knowledge base
  representation.
\newblock In: ECCV.
\newblock (2014)  408--424

\bibitem{akata2015evaluation}
Akata, Z., Reed, S., Walter, D., Lee, H., Schiele, B.:
\newblock Evaluation of output embeddings for fine-grained image
  classification.
\newblock In: CVPR. (2015)  2927--2936

\bibitem{deng2012hedging}
Deng, J., Krause, J., Berg, A.C., Fei-Fei, L.:
\newblock Hedging your bets: Optimizing accuracy-specificity trade-offs in
  large scale visual recognition.
\newblock In: CVPR. (2012)  3450--3457

\bibitem{chao2015mining}
Chao, Y.W., Wang, Z., Mihalcea, R., Deng, J.:
\newblock Mining semantic affordances of visual object categories.
\newblock In: CVPR. (2015)  4259--4267

\bibitem{castellini2011using}
Castellini, C., Tommasi, T., Noceti, N., Odone, F., Caputo, B.:
\newblock Using object affordances to improve object recognition.
\newblock Autonomous Mental Development \textbf{3}(3) (2011)  207--215

\bibitem{zhu2015understanding}
Zhu, Y., Zhao, Y., Chun~Zhu, S.:
\newblock Understanding tools: Task-oriented object modeling, learning and
  recognition.
\newblock In: CVPR. (2015)  2855--2864

\bibitem{koppula2016anticipating}
Koppula, H.S., Saxena, A.:
\newblock Anticipating human activities using object affordances for reactive
  robotic response.
\newblock PAMI \textbf{38}(1) (2016)  14--29

\bibitem{kjellstrom2011visual}
Kjellstr{\"o}m, H., Romero, J., Kragi{\'c}, D.:
\newblock Visual object-action recognition: Inferring object affordances from
  human demonstration.
\newblock CVIU \textbf{115}(1) (2011)  81--90

\bibitem{lenz2015deep}
Lenz, I., Lee, H., Saxena, A.:
\newblock Deep learning for detecting robotic grasps.
\newblock IJRR \textbf{34}(4-5) (2015)  705--724

\bibitem{desai2013predicting}
Desai, C., Ramanan, D.:
\newblock Predicting functional regions on objects.
\newblock In: CVPR Workshops. (2013)  968--975

\bibitem{song2015learning}
Song, H.O., Fritz, M., Goehring, D., Darrell, T.:
\newblock Learning to detect visual grasp affordance.
\newblock (2015)

\bibitem{grabner2011makes}
Grabner, H., Gall, J., Van~Gool, L.:
\newblock What makes a chair a chair?
\newblock In: CVPR. (2011)  1529--1536

\bibitem{Jiang_2013_CVPR}
Jiang, Y., Koppula, H., Saxena, A.:
\newblock Hallucinated humans as the hidden context for labeling 3d scenes.
\newblock In: CVPR. (June)

\bibitem{koppula2014physically}
Koppula, H.S., Saxena, A.:
\newblock Physically grounded spatio-temporal object affordances.
\newblock In: ECCV.
\newblock (2014)  831--847

\bibitem{vezhnevets2010towards}
Vezhnevets, A., Buhmann, J.M.:
\newblock Towards weakly supervised semantic segmentation by means of multiple
  instance and multitask learning.
\newblock In: CVPR. (2010)  3249--3256

\bibitem{vezhnevets2011weakly}
Vezhnevets, A., Ferrari, V., Buhmann, J.M.:
\newblock Weakly supervised semantic segmentation with a multi-image model.
\newblock In: ICCV. (2011)  643--650

\bibitem{vezhnevets2012weakly}
Vezhnevets, A., Ferrari, V., Buhmann, J.M.:
\newblock Weakly supervised structured output learning for semantic
  segmentation.
\newblock In: CVPR. (2012)  845--852

\bibitem{xu2014tell}
Xu, J., Schwing, A., Urtasun, R.:
\newblock Tell me what you see and i will show you where it is.
\newblock In: CVPR. (2014)  3190--3197

\bibitem{zhang2015weakly}
Zhang, W., Zeng, S., Wang, D., Xue, X.:
\newblock Weakly supervised semantic segmentation for social images.
\newblock In: CVPR. (2015)  2718--2726

\bibitem{chen2014semantic}
Chen, L.C., Papandreou, G., Kokkinos, I., Murphy, K., Yuille, A.L.:
\newblock Semantic image segmentation with deep convolutional nets and fully
  connected crfs.
\newblock ICLR (2015)

\bibitem{pathak2015constrained}
Pathak, D., Krahenbuhl, P., Darrell, T.:
\newblock Constrained convolutional neural networks for weakly supervised
  segmentation.
\newblock In: ICCV. (2015)  1796--1804

\bibitem{simonyan2014very}
Simonyan, K., Zisserman, A.:
\newblock Very deep convolutional networks for large-scale image recognition.
\newblock arXiv preprint arXiv:1409.1556 (2014)

\bibitem{hariharan2015hypercolumns}
Hariharan, B., Arbel{\'a}ez, P., Girshick, R., Malik, J.:
\newblock Hypercolumns for object segmentation and fine-grained localization.
\newblock In: CVPR. (2015)  447--456

\end{thebibliography}
\end{document}